\documentclass[11pt,a4paper]{article}
\usepackage[hyperref]{acl2021}
\usepackage{times}
\usepackage{latexsym}

\usepackage{microtype}
\usepackage{amsmath}
\usepackage{multirow}
\usepackage{graphicx}
\usepackage{booktabs}
\usepackage{subfigure}
\usepackage{arydshln}
\usepackage{amssymb}
\definecolor{diff}{RGB}{255,50,50}
\definecolor{diffword}{RGB}{252,80,80}
\usepackage[german]{babel}

\aclfinalcopy 

\newcommand{\diffword}[1]{{\color{diffword} #1}}

\newcommand{\tabincell}[2]{\begin{tabular}{@{}#1@{}}#2\end{tabular}}  

\title{Difficulty-Aware Machine Translation Evaluation}

\author{Runzhe Zhan\thanks{~~Equal contribution}~~~
        Xuebo Liu$\footnotemark[1]$~~~
        Derek F. Wong\thanks{~~Corresponding author}~~~
        Lidia S. Chao \\
  NLP$^2$CT Lab, Department of Computer and Information Science, 
  University of Macau\\
  \texttt{nlp2ct.\{runzhe,xuebo\}@gmail.com, \{derekfw,lidiasc\}@um.edu.mo} \\
}

\date{}

\begin{document}
\maketitle
\begin{abstract}
The high-quality translation results produced by machine translation (MT) systems still pose a huge challenge for automatic evaluation. 
Current MT evaluation pays the same attention to each sentence component, while the questions of real-world examinations (e.g., university examinations) have different difficulties and weightings. 
In this paper, we propose a novel {\it difficulty-aware MT evaluation} metric, expanding the evaluation dimension by taking translation difficulty into consideration.
A translation that fails to be predicted by most MT systems will be treated as a difficult one and assigned a large weight in the final score function, and conversely.
Experimental results on the WMT19 English$\leftrightarrow$German Metrics shared tasks show that our proposed method outperforms commonly-used MT metrics in terms of human correlation.
In particular, our proposed method performs well even when all the MT systems are very competitive, which is when most existing metrics fail to distinguish between them.
The source code is freely available at \texttt{\href{https://github.com/NLP2CT/Difficulty-Aware-MT-Evaluation}{https://github.com/NLP2CT /Difficulty-Aware-MT-Evaluation}}.
\end{abstract}

\section{Introduction}
The human labor needed to evaluate machine translation (MT) evaluation is expensive.
To alleviate this, various automatic evaluation metrics are continuously being introduced to correlate with human judgements. 
Unfortunately, cutting-edge MT systems are too close in performance and generation style for such metrics to rank systems.
Even for a metric whose correlation is reliable in most cases, empirical research has shown that it poorly correlates with human ratings when evaluating competitive systems \cite{ma-etal-2019-results, mathur-etal-2020-tangled}, limiting the development of MT systems.

Current MT evaluation still faces the challenge of how to better evaluate the overlap between the reference and the model hypothesis taking into consideration \textit{adequacy} and \textit{fluency}, where all the evaluation units are treated the same, i.e., all the matching scores have an equal weighting.
However, in real-world examinations, the questions vary in their difficulty.
Those questions which are easily answered by most subjects tend to have low weightings, while those which are hard to answer have high weightings.
A subject who is able to solve the more difficult questions can receive a high final score and gain a better ranking. 
MT evaluation is also a kind of examination.
For bridging the gap between human examination and MT evaluation, it is advisable to incorporate a \textit{difficulty} dimension into the MT evaluation metric.

In this paper, we take translation difficulty into account in MT evaluation and test the effectiveness on a representative MT metric BERTScore~\citep{zhang2019bertscore} to verify the feasibility. More specifically, the difficulty is first determined across the systems with the help of pairwise similarity, and then exploited as the weight in the final score function for distinguishing the contribution of different sub-units.
Experimental results on the WMT19 English$\leftrightarrow$German evaluation task show that difficulty-aware BERTScore has a better correlation than do the existing metrics.
Moreover, it agrees very well with the human rankings when evaluating competitive systems.

\begin{figure*}[t]
    \centering
    \includegraphics[width=16cm]{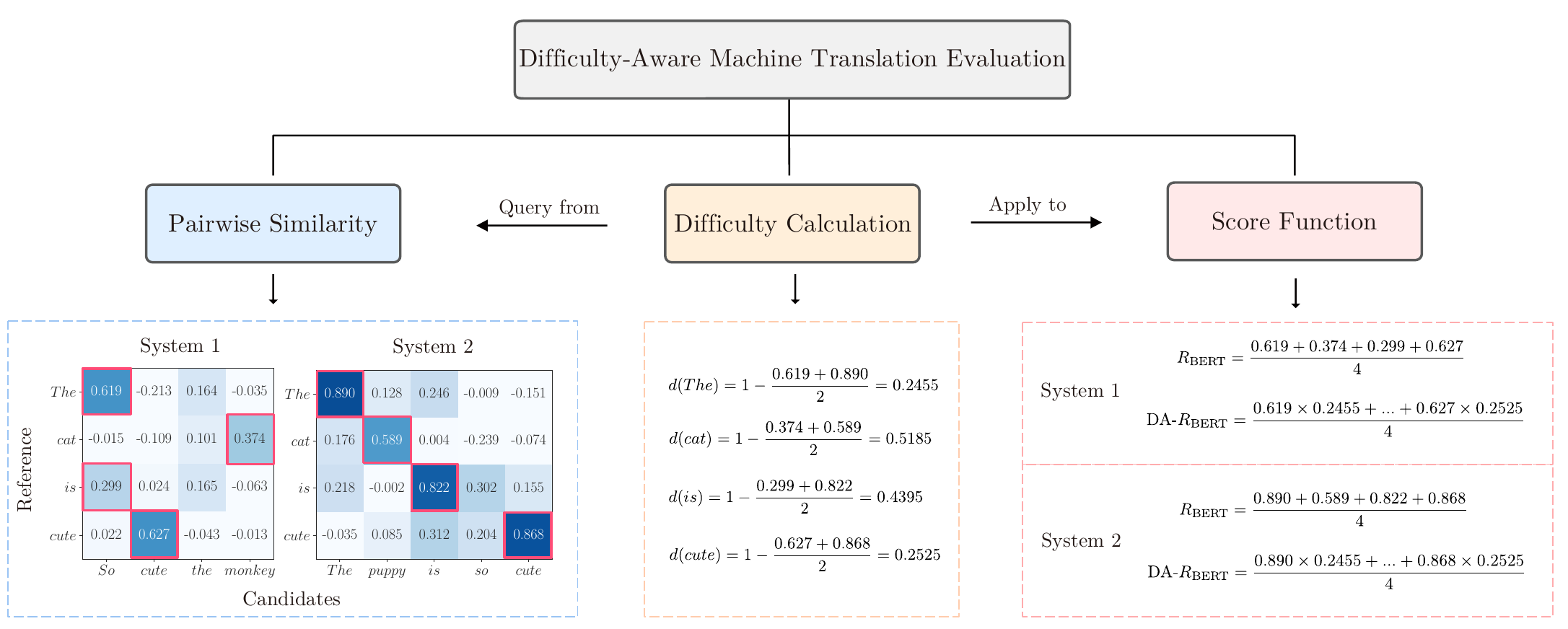}
    \caption{Illustration of combining difficulty weight with BERTScore. $R_\mathrm{BERT}$ denotes the vanilla recall-based BERTScore while $\mathrm{DA}\text{-}R_\mathrm{BERT}$ denotes the score augmented with translation difficulty.}
    \label{fig:fig_arch}
\end{figure*}

\section{Related Work}
The existing MT evaluation metrics can be categorized into the following types according to their underlying matching sub-units: $n$-gram based \cite{papineni-etal-2002-bleu,10.5555/1289189.1289273,lin-och-2004-automatic,han-etal-2012-lepor,popovic-2015-chrf}, edit-distance based \cite{Snover06astudy,leusch-etal-2006-cder}, alignment-based \cite{banerjee-lavie-2005-meteor}, embedding-based \cite{zhang2019bertscore,chow-etal-2019-wmdo,lo-2019-yisi} and end-to-end based \cite{sellam-etal-2020-bleurt}. BLEU \cite{papineni-etal-2002-bleu} is widely used as a vital criterion in the comparison of MT system performance but its reliability has been doubted on entering neural machine translation age \cite{shterionov2018human, mathur-etal-2020-tangled}. Due to the fact that BLEU and its variants only assess surface linguistic features, some metrics leveraging contextual embedding and end-to-end training bring semantic information into the evaluation, which further improves the correlation with human judgement. Among them, BERTScore \cite{zhang2019bertscore} has achieved a remarkable performance across MT evaluation benchmarks balancing speed and correlation. 
In this paper, we choose BERTScore as our testbed.

\section{Our Proposed Method}

\subsection{Motivation}
    In real-world examinations, the questions are empirically divided into various levels of difficulty.
    Since the difficulty varies from question to question, the corresponding role a question plays in the evaluation does also. 
    Simple question, which can be answered by most of the subjects, usually receive of a low weighting.
    But a difficult question, which has more discriminative power, can only be answered by a small number of good subjects, and thus receives a higher weighting.
    
    Motivated by this evaluation mechanism, we measure difficulty of a translation by viewing the MT systems and sub-units of the sentence as the subjects and questions, respectively.
    From this perspective, the impact of the sentence-level sub-units on the evaluation results supported a differentiation.
    Those sub-units that may be incorrectly translated by most systems (e.g., polysemy) should have a higher weight in the assessment, while easier-to-translate sub-units (e.g., the definite article) should receive less weight.

\begin{table*}[t]
\centering
\scalebox{0.85}{
\begin{tabular}{lcccccccccccc}
\toprule
\multicolumn{1}{c}{\multirow{2}{*}{\textbf{Metric}}} & \multicolumn{3}{c}{\textbf{En$\rightarrow$De (All) }}   & \multicolumn{3}{c}{\textbf{En$\rightarrow$De (Top~30\%)}} & \multicolumn{3}{c}{\bf De$\rightarrow$En (All)} & \multicolumn{3}{c}{\bf De$\rightarrow$En (Top~30\%) }                   \\ 
\cmidrule(r){2-4} \cmidrule(r){5-7} \cmidrule(r){8-10} \cmidrule(r){11-13}
\multicolumn{1}{c}{}                        & $|r|$         & $|\tau|$         & $|\rho|$        & $|r|$         & $|\tau|$         & $|\rho|$  & $|r|$         & $|\tau|$         & $|\rho|$        & $|r|$         & $|\tau|$         & $|\rho|$      \\ \midrule
\bf BLEU                    & 0.952          & 0.703          & 0.873          & 0.460          & 0.200          & 0.143          & 0.888          & 0.622          & 0.781          & 0.808          & 0.548          & 0.632          \\
\bf TER                     & 0.982          & 0.711          & 0.873          & 0.598          & 0.333          & 0.486          & 0.797          & 0.504          & 0.675          & \bf 0.883          & 0.548          & 0.632          \\
\bf METEOR                  & 0.985          & 0.746          & 0.904          & 0.065          & 0.067          & 0.143          & 0.886          & 0.605          & 0.792          & 0.632          & 0.548          & 0.632          \\
\bf BERTScore                 & 0.990          & 0.772          & 0.920          & 0.204          & 0.067          & 0.143          & 0.949          & 0.756          & 0.890          & 0.271 & 0.183          & 0.316          \\ \hdashline
\bf DA-BERTScore             & \textbf{0.991} & \textbf{0.798} & \textbf{0.930} & \textbf{0.974} & \textbf{0.733} & \textbf{0.886} & \textbf{0.951} & \textbf{0.807} & \textbf{0.932} & 0.693          & \textbf{0.548} & \textbf{0.632} \\
\bottomrule
\end{tabular}}
\caption{\label{tab:main_result}Absolute correlations with system-level human judgments on WMT19 metrics shared task. For each metric, higher values are better. Difficulty-aware BERTScore consistently outperforms vanilla BERTScore across different evaluation metrics and translation directions, especially when the evaluated systems are very competitive (i.e., evaluating on the top 30\% systems).} 
\end{table*}
   
\subsection{Difficulty-Aware BERTScore}
In this part, we aim to answer two questions: 1) how to automatically collect the translation difficulty from BERTScore; and 2) how to integrate the difficulty into the score function.
Figure \ref{fig:fig_arch} presents an overall illustration.

    \paragraph{Pairwise Similarity}
    Traditional $n$-gram overlap cannot extract semantic similarity, word embedding provides a means of quantifying the degree of overlap, which allows obtaining more accurate difficulty information. Since BERT is a strong language model, it can be utilized as a contextual embedding $\mathbf{O}_{\mathrm{_{BERT}}}$ (i.e., the output of BERT) for obtaining the representations of the reference $\mathbf{t}$ and the hypothesis $\mathbf{h}$. 
    Given a specific hypothesis token $h$ and reference token $t$, the similarity score $\mathrm{sim}(t,h)$ is computed as follows:
    \begin{equation}
        \mathrm{sim}(t,h) = \frac{\mathbf{O}_{\mathrm{_{BERT}}}(t)^\mathsf{T}\mathbf{O}_{\mathrm{_{BERT}}}(h)}{\left\lVert\mathbf{O}_{\mathrm{_{BERT}}}(t)\right\rVert \cdot \left\lVert\mathbf{O}_{\mathrm{_{BERT}}}(h)\right\rVert}
    \end{equation}
    Subsequently, a similarity matrix is constructed by pairwise calculating the token similarity. Then the token-level matching score is obtained by greedily searching for the maximal similarity in the matrix, which will be further taken into account in sentence-level score aggregation. 
    
    \paragraph{Difficulty Calculation} The calculation of difficulty can be tailored for different metrics based on the overlap matching score. In this case, BERTScore evaluates the token-level overlap status by the pairwise semantic similarity, thus the token-level similarity is viewed as the bedrock of difficulty calculation. For instance, if one token (like ``cat'') in the reference may only find identical or synonymous substitutions in a few MT system outputs, then the corresponding translation difficulty weight ought to be larger than for other reference tokens, which further indicates that it is more valuable for evaluating the translation capability. 
    Combined with BERTScore mechanism, it is implemented by averaging the token similarities across systems. Given $K$ systems and their corresponding generated hypotheses $\mathbf{h}_1, \mathbf{h}_2, ..., \mathbf{h}_K$, the difficulty of a specific token $t$ in the reference $\mathbf{t}$ is formulated as
    	\begin{equation}
    		d(t) = 1 - \frac{\sum_{k=1}^{K} \max_{h \in \mathbf{h}_{k}} \mathrm{sim}(t, h)}{K}
    	\end{equation}

    An example is shown in Figure \ref{fig:fig_arch}: the entity ``cat'' is improperly translated to ``monkey'' and ``puppy'', resulting in a lower pairwise similarity of the token ``cat'', which indicates higher translation difficulty. Therefore, by incorporating the translation difficulty into the evaluation process, the token ``cat'' is more contributive while the other words like ``cute'' are less important in the overall score.

\paragraph{Score Function}
Due to the fact that the translation generated by a current NMT model is fluent enough but not adequate yet, $F$-score which takes into account the \textit{Precision} and \textit{Recall}, is more appropriate to aggregate the matching scores, instead of only considering precision.
We thus follow vanilla BERTScore in using F-score as the final score.
The proposed method directly assigns difficulty weights to the counterpart of the similarity score {\bf without any hyperparameter}:
 \begin{equation} \label{da-recall}
    	\mathrm{DA}\text{-}R_\mathrm{BERT} = \frac{1}{|\mathbf{t}|} \sum_{t \in \mathbf{t}} d(t) \max_{h \in \mathbf{h}} \mathrm{sim}(t, h)
    \end{equation}
\begin{equation} \label{da-prec}
    	\mathrm{DA}\text{-}P_\mathrm{BERT} = \frac{1}{|\mathbf{h}|} \sum_{h \in \mathbf{h}} d(h) \max_{t \in \mathbf{t}} \mathrm{sim}(t, h)
\end{equation}
\begin{equation}\label{da-Fbert}
    	\mathrm{DA}\text{-}F_\mathrm{BERT} = 2 \cdot \frac{\mathrm{DA}\text{-}R_\mathrm{BERT} \cdot \mathrm{DA}\text{-}P_\mathrm{BERT}}{\mathrm{DA}\text{-}R_\mathrm{BERT} + \mathrm{DA}\text{-}P_\mathrm{BERT}}
\end{equation}
\noindent For any $h \notin \mathbf{t}$, we simply let $d(h)=1$, i.e., retaining the original calculation. The motivation is that the human assessor keeps their initial matching judgement if the test taker produces a unique but reasonable alternative answer. 
We regard $\mathrm{DA}\text{-}F_\mathrm{BERT}$ as the DA-BERTScore in the following part.

There are many variants of our proposed method: 1) designing more elaborate difficulty function \cite{liu-etal-2020-norm, zhan-etal-2021-metacl}; 2) applying a smoothing function to the difficulty distribution; and 3) using other kinds of $F$-score, e.g., $F_{0.5}$-score.
The aim of this paper is not to explore this whole space but simply to show that a straightforward implementation works well for MT evaluation.

\section{Experiments}

\begin{table*}[t]
\centering
\scalebox{0.85}{
\begin{tabular}{ccccccc}
\toprule
\textbf{SYSTEM} & \textbf{BLEU~$\uparrow$}       & \textbf{TER~$\downarrow$}        & \textbf{METEOR~$\uparrow$}     & \textbf{BERTScore~$\uparrow$}  & \textbf{DA-BERTScore~$\uparrow$}  & \textbf{HUMAN~$\uparrow$}      \\ \midrule
\bf Facebook.6862   & 0.4364 ($\Downarrow$5)         & 0.4692 ($\Downarrow$5)         & 0.6077 ($\Downarrow$3)         & 0.7219 ($\Downarrow$4)         & \textbf{0.1555 ($\checkmark$0)}  & \textbf{0.347} \\
\bf Microsoft.sd.6974 & 0.4477 ($\Downarrow$1)         & 0.4583 ($\Downarrow$1)         & 0.6056 ($\Downarrow$3)         & 0.7263 ($\checkmark$0)          & 0.1539 ($\Downarrow$1) & 0.311 \\
\bf Microsoft.dl.6808 & 0.4483 ($\Uparrow$1)          & 0.4591 ($\Downarrow$1)          & 0.6132 ($\Uparrow$1)         & 0.7260 ($\checkmark$0)          & 0.1544 ($\Uparrow$1)  & 0.296\\
\bf MSRA.6926 & \textbf{0.4603 ($\Uparrow$3)} & \textbf{0.4504 ($\Uparrow$3)} & \textbf{0.6187 ($\Uparrow$3)}         & \textbf{0.7267 ($\Uparrow$3)} & 0.1525 ($\checkmark$0)        & 0.214  \\
\bf UCAM.6731 & 0.4413 ($\checkmark$0)          & 0.4636 ($\checkmark$0)         & 0.6047 ($\Downarrow$1)          & 0.7190 ($\Downarrow$1)          & 0.1519 ($\Downarrow$1)  & 0.213 \\
\bf NEU.6763    & 0.4460 ($\Uparrow$2)          & 0.4563 ($\Uparrow$4)          & 0.6083 ($\Uparrow$3)          & 0.7229 ($\Uparrow$2)          & 0.1521 ($\Uparrow$1)  & 0.208      \\ \midrule 
\bf $\mathrm{sum}(|\triangle_\mathrm{Rank}|)$   & 12          &  14         & 14  &   10        &  \bf 4  & 0 \\ 
\bottomrule   
\end{tabular}}
\caption{\label{tab:ranking-study}Agreement of system ranking with human judgement on the top 30\% systems (k=6) of WMT19 En$\rightarrow$De Metrics task. $\Uparrow$/$\Downarrow$ denotes that the rank given by the evaluation metric is higher/lower than human judgement, and $\checkmark$ denotes that the given rank is equal to human ranking. DA-BERTScore successfully ranks the best system that the other metrics failed. Besides, it also shows the lowest rank difference.}
\end{table*}

\paragraph{Data} The WMT19 English$\leftrightarrow$German (En$\leftrightarrow$De) evaluation tasks are challenging due to the large discrepancy between human and automated assessments in terms of reporting the best system \cite{bojar-etal-2018-findings,barrault-etal-2019-findings,freitag-etal-2020-bleu}. 
To sufficiently validate the effectiveness of our approach, we choose these tasks as our evaluation subjects. 
There are 22 systems for En$\rightarrow$De and 16 for De$\rightarrow$En. Each system has its corresponding human assessment results. 
The experiments were centered on the correlation with system-level human ratings.

\paragraph{Comparing Metrics} In order to compare with the metrics that have different underlying evaluation mechanism, four representative metrics: BLEU \cite{papineni-etal-2002-bleu}, TER \cite{Snover06astudy}, METEOR \cite{banerjee-lavie-2005-meteor, denkowski:lavie:meteor-wmt:2014}, BERTScore \cite{zhang2019bertscore}, which are correspondingly driven by $n$-gram, edit distance, word alignment and embedding similarity, are involved in the comparison experiments without losing popularity. For ensuring reproducibility, the original\footnote{https://www.cs.cmu.edu/~alavie/METEOR/index.html}\footnote{https://github.com/Tiiiger/bert\_score} and widely used implementation\footnote{https://github.com/mjpost/sacrebleu} was used in the experiments. 

\begin{figure}[t]
    \centering
    \includegraphics[scale=0.49]{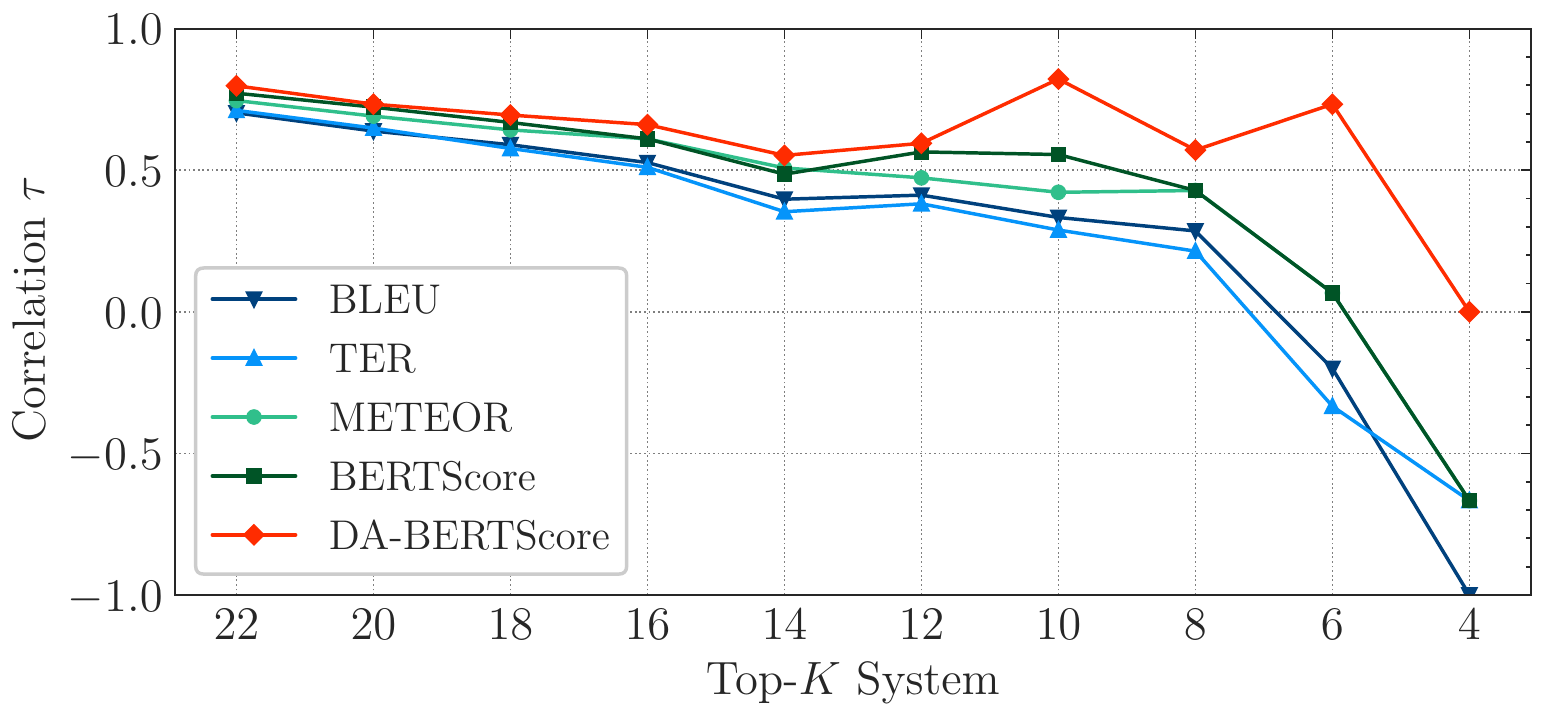} 
    \caption{Effect of top-$K$ systems in the En$\rightarrow$De evaluation. DA-BERTScore is highly correlated with human judgment for different values of $K$, especially when all the systems are competitive (i.e., $K\leq$10).}
    \label{fig:topk}
\end{figure}

\paragraph{Main Results} Following the correlation criterion adopted by the WMT official organization, Pearson’s correlation $r$ is used for validating the system-level correlation with human ratings. In addition, two rank-correlations Spearman's $\rho$ and original Kendall's $\tau$ are also used to examine the agreement with human ranking, as has been done in recent research \cite{freitag-etal-2020-bleu}. 
Table \ref{tab:main_result} lists the results.
DA-BERTScore achieves competitive correlation results and further improves the correlation of BERTScore.
In addition to the results on all systems, we also present the results on the top 30\% systems where the calculated difficulty is more reliable and our approach should be more effective.
The result confirms our intuition that DA-BERTScore can significantly improve the correlations under the competitive scenario, e.g., improving the $|r|$ score from 0.204 to 0.974 on En$\rightarrow$De and 0.271 to 0.693 on De$\rightarrow$En.

\begin{table*}[t!]
\centering
\scalebox{0.8}{
\begin{tabular}{llll}
\toprule
& \bf BERTS. & \bf +DA & \multicolumn{1}{c}{\bf Sentence} \\
\midrule
\bf Src & - & - & ``I'm standing \diffword{\bf{right here}} in front of you,'' one woman said. \\
\bf Ref &- & - &  \glqq Ich stehe \colorbox{diff!30}{genau} hier vor Ihnen \grqq, sagte eine Frau. \\
\hdashline
\bf MSRA & \bf 0.9656 &  0.0924 &  \glqq Ich stehe \diffword{\bf{hier vor}} Ihnen \grqq, sagte eine Frau.  \\
\bf Facebook  & 0.9591 & \bf 0.1092 & \glqq Ich stehe \diffword{\bf{hier direkt vor}} Ihnen \grqq, sagte eine Frau.  \\
\hline\hline
\bf Src  & - & - & France has more than 1,000 \diffword{\bf{troops on the ground}} in the war-wracked country.  \\
\bf Ref & - & - & Frankreich hat über 1.000 \colorbox{diff!30}{Bodensoldaten} in dem kriegszerstörten Land im Einsatz.  \\
\hdashline
\bf MSRA & \bf 0.6885 & 0.2123 & Frankreich hat mehr als 1.000 \diffword{\bf{Soldaten vor Ort}} in dem kriegsgeplagten Land. \\
\bf Facebook & 0.6772 & \bf 0.2414 & \tabincell{l}{Frankreich hat mehr als 1000 \diffword{\bf{Soldaten am Boden}} in dem kriegsgeplagten Land} stationiert. \\
\bottomrule
\end{tabular}}
\caption{\label{tab:token_case} Examples from the En$\rightarrow$De evaluation. BERTS. denotes BERTScore. \colorbox{diff!30}{Words} indicate the difficult translations given by our approach on the top 30\% systems. DA-BERTScores are more in line with human judgements.} 
\end{table*}

\paragraph{Effect of Top-$K$ Systems}
Figure \ref{fig:topk} compares the Kendall's correlation variation of the top-$K$ systems.
Echoing previous research, the vast majority of metrics fail to correlate with human ranking and even perform negative correlation when $K$ is lower than $6$, meaning that the current metrics are ineffective when facing competitive systems. 
With the help of difficulty weights, the degradation in the correlation is alleviated, e.g., improving $\tau$ score from 0.07 to 0.73 for BERTScore ($K=6$). 
These results indicate the effectiveness of our approach, establishing the necessity for adding difficulty.

\paragraph{Case Study of Ranking}
Table \ref{tab:ranking-study} presents a case study on the En$\rightarrow$De task.
Existing metrics consistently select MSRA's system as the best system, which shows a large divergence from human judgement.
DA-BERTScore ranks it the same as human (4th) because most of its translations have low difficulty, thus lower weights are applied in the scores.
Encouragingly, DA-BERTScore ranks Facebook's system as the best one, which implies that it overcomes more challenging translation difficulties.
This testifies to the importance and effectiveness of considering translation difficulty in MT evaluation. 

\paragraph{Case Study of Token-Level Difficulty}
Table~\ref{tab:token_case} presents two cases, illustrating that our proposed difficulty-aware method successfully identifies the omission errors ignored by BERTScore.
In the first case, the Facebook's system correctly translates the token ``right'', and in the second case, uses the substitute ``Soldaten am Boden'' which is lexically similar to the ground-truth token ``Bodensoldaten''.
Although the MSRA's system suffers word omissions in the two cases, its hypotheses receive the higher ranking given by BERTScore, which is inconsistent with human judgements.
The reason might be that the semantic of the hypothesis is highly close to the reference, thus the slight lexical difference is hard to be found when calculating the similarity score.
By distinguishing the difficulty of the reference tokens, DA-BERTScore successfully makes the evaluation focus on the difficult parts, and eventually correct the score of the Facebook's system, thus giving the right rankings.

\begin{figure}[t]
    \centering
    \includegraphics[scale=0.49]{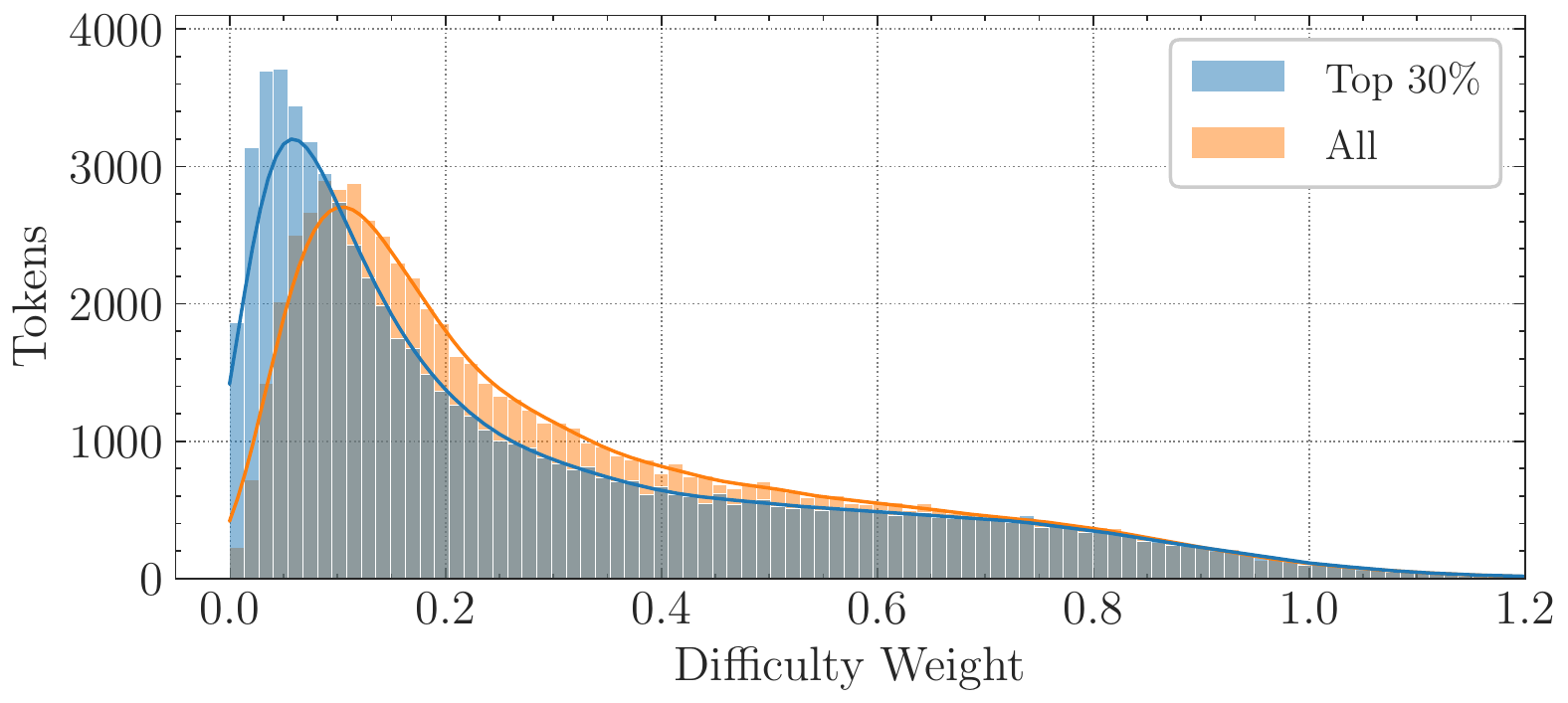}
    \caption{Distribution of token-level difficulty weights extracted from the En$\rightarrow$De evaluation.}
    \label{fig:diff_distribution}
\end{figure}
\paragraph{Distribution of Difficulty Weights}
The difficulty weights can reflect the translation ability of a group of MT systems. 
If the systems in a group are of higher translation ability, the calculated difficulty weights will be smaller.
Starting from this intuition, we visualize the distribution of difficulty weights as shown in Figure \ref{fig:diff_distribution}. 
Clearly, we can see that the difficulty weights are centrally distributed at lower values, indicating that most of the tokens can be correctly translated by all the MT systems.
For the difficulty weights calculated on the top 30\% systems, the whole distribution skews to zero since these competitive systems have better translation ability and thus most of the translations are easy for them.
This confirms that the difficulty weight produced by our approach is reasonable.

\section{Conclusion and Future Work}
This paper introduces the conception of difficulty into machine translation evaluation, and verifies our assumption with a representative metric BERTScore.
Experimental results on the WMT19 English$\leftrightarrow$German metric tasks show that our approach achieves a remarkable correlation with human assessment, especially for evaluating competitive systems, revealing the importance of incorporating difficulty into machine translation evaluation.
Further analyses show that our proposed difficulty-aware BERTScore can strengthen the evaluation of word omission problems and generate reasonable distributions of difficulty weights.

Future works include: 1) optimizing the difficulty calculation \cite{zhan2021variance}; 2) applying to other MT metrics; and 3) testing on other generation tasks, e.g., text summarization. 

\section*{Acknowledgement} 
This work was supported in part by the Science and Technology Development Fund, Macau SAR (Grant No. 0101/2019/A2), and the Multi-year Research Grant from the University of Macau (Grant No. MYRG2020-00054-FST). We thank the anonymous reviewers for their insightful comments.

\bibliographystyle{acl_natbib}
\bibliography{anthology,acl2021}

\end{document}